\documentclass[sigconf]{acmart}
\usepackage{tabularx}
\usepackage{enumitem}
\usepackage{url}
\AtBeginDocument{%
  }

\copyrightyear{2026}
\acmYear{2026}
\setcopyright{cc}
\setcctype{by}
\acmConference[UIST '26]{The 39th Annual ACM Symposium on User Interface Software and Technology}{November 02--05, 2026}{Detroit, MI, USA}
\acmBooktitle{The 39th Annual ACM Symposium on User Interface Software and Technology (UIST '26), November 02--05, 2026, Detroit, MI, USA}
\acmDOI{10.1145/3830398.3830553}
\acmISBN{979-8-4007-2856-3/2026/11}
\begin{document}

\title{TransfHAR: Self-Supervised Wrist Representations for On-Demand Activity Recognition}

\author{Aidan Bradshaw}
\orcid{0009-0007-1731-8786}
\email{aidanbradshaw2025@u.northwestern.edu}
\affiliation{%
  \institution{Northwestern University}
  \city{Evanston}
  \state{Illinois}
  \country{USA}
}

\author{Riku Arakawa}
\orcid{0000-0001-7868-4754}
\email{rarakawa@cs.cmu.edu}
\affiliation{%
  \institution{Carnegie Mellon University}
  \city{Pittsburgh}
  \state{Pennsylvania}
  \country{USA}
}

\author{Xin Liu}
\orcid{0000-0002-9279-5386}
\email{xliu0@cs.washington.edu}
\affiliation{%
  \institution{Google}
  \city{Seattle}
  \state{Washington}
  \country{USA}
}

\author{Karan Ahuja}
\orcid{0000-0003-2497-0530}
\email{kahuja@northwestern.edu}
\affiliation{%
  \institution{Northwestern University}
  \city{Evanston}
  \state{Illinois}
  \country{USA}
}

\renewcommand{\shortauthors}{Bradshaw et al.}

\begin{teaserfigure}
    \centering
    \includegraphics[width=\textwidth]{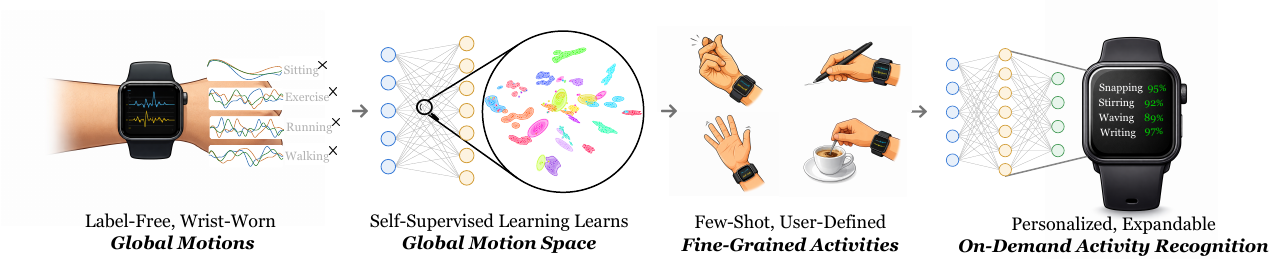}
    \caption{TransfHAR pretrains a ViT-1D encoder with self-supervised masked autoencoding on IMU data of global wrist motions (running, walking, sitting; left). The learned representations transfer to fine-grained, user-defined activities unseen in pretraining (snapping, stirring, writing; center). At deployment, a lightweight linear probe on the frozen encoder learns from a few on-watch demonstrations, enabling personalized, expandable on-demand recognition (right).}
    \Description{A four-stage pipeline read left to right. The first stage shows a wrist wearing a smartwatch beside four motion traces labeled sitting, exercise, running, and walking, each label struck through to indicate that class labels are discarded. The second stage shows a neural network schematic with a magnified inset of a dense multicolored embedding scatter, standing for the learned global motion space. The third stage shows four photographs of a watch-wearing hand performing fine-grained activities including finger snapping, writing with a pen, an open waving palm, and stirring a cup of coffee. The final stage shows a second network whose green output layer feeds a watch screen listing snapping, stirring, waving, and writing with confidence values in the high eighties and nineties.}
    \label{fig:teaser}
\end{teaserfigure}

\begin{abstract}
Fine-grained wrist activity recognition can support applications such as procedural step guidance and context-aware assistance, yet acquiring labeled data for every new task, user, and activity granularity remains a bottleneck. We present TransfHAR, a self-supervised wrist IMU framework for on-demand, fine-grained activity recognition by learning transferable motion priors from global, unlabeled activities. We show that self-supervised pretraining on coarse wrist IMU activities (e.g., sitting, walking, exercise) learns motion structure rich enough to transfer to fine-grained manipulative, gestural, and procedural activities (e.g., snapping, stirring, waving) that are absent from pretraining. We implement TransfHAR as a real-time smartwatch application that lets users define and expand their own activity set for personalized recognition from only a few demonstrations. Across three offline cross-dataset evaluations, TransfHAR matches or exceeds fully supervised baselines that use complete label sets with equal or additional sensor channels, by 6.2 balanced-accuracy points on average. In an in-lab study with 10 participants each performing seven novel wrist activities, TransfHAR reaches 86.7\% balanced accuracy across participants with five examples per class and 90.4\% when updated from a single one-minute recording per class. These results indicate that broad self-supervised wrist pretraining provides an effective foundation for on-demand fine-grained activity recognition.

\end{abstract}


\maketitle

\section{Introduction}
Everyone has different habits, shortcuts, and mannerisms that shape how they move through their daily lives. This has inspired wrist-worn activity recognition systems that can classify a breadth of human activities directly on a smartwatch, ranging from fine hand gestures~\cite{Serendipity, xu2022enabling, LaputFineGrainedHandActivity} to activities of daily living~\cite{Bhattacharya2022UTWatch, yeon2025watchhar}, enabling new applications that support habitual, routine, and diverse context-aware interactions~\cite{arakawa2024prismobserver,DBLP:journals/imwut/ArakawaLG24,DBLP:journals/pacmhci/ArakawaNY25}. These types of systems are typically designed around pre-defined activity sets, assumptions about persisting activity granularity, and models trained to generalize across users rather than adapt to individual routines~\cite{CrossHAR, Bhattacharya2022UTWatch,DBLP:conf/uist/ZhouAAG25}. In practice, however, benchmark-driven activity classes rarely match the activities users actually need. Current models are trained on fixed label vocabularies that cannot be extended after deployment, and adding new activities requires collecting fresh labeled data and retraining from scratch~\cite{Mollyn2022SAMOSA, yeon2025watchhar, LIMUBERT}. A barista may want to track the steps of their pour-over routine, a rehabilitation patient may need to monitor exercises prescribed by their therapist, or a developer may want gesture shortcuts tailored to their workflow. Most of these personalized activities do not exist in any pre-existing training set, and each is defined by the user's own context, tools, and goals. To maximize utility for end users, the underlying models should be built to optimize for flexibility across motion types~\cite{Lockhart2014Limitations} and support the self-definition of people's own activities and mannerisms.

The direct approach to this problem, collecting a labeled dataset for each new user-defined activity and training a task-specific classifier, does not scale to the diversity of activities people actually care about, and training from scratch with limited examples is ineffective. Some prior work addresses this through gesture customization or incremental adaptation~\cite{xu2022enabling, uWave}, but these approaches typically extend a fixed gesture vocabulary within a single interaction domain rather than supporting arbitrary new activities across motion and behavior types. The underlying challenge is that fine-grained manipulative, gestural, and procedural activities, the kinds most relevant to personalization, are scarce in public wrist IMU datasets, which are overrepresented by coarse locomotion, posture, and exercise data~\cite{Chan2024CAPTURE24, Ciliberto2021Opportunity, Morris2014RecoFit}. However, since these datasets are collected from the same wrist-worn IMU modality, they still capture many of the underlying motion primitives that reappear in fine-grained personalized activities, even when those subtasks are not explicitly labeled. In this case, self-supervised learning could offer a promising path forward, as prior work in wearable sensing and time-series learning has shown that label-free objectives can learn transferable, within-domain motion manifolds from large amounts of unlabeled data while reducing dependence on expensive manual annotation~\cite{Hang2024SelfSupervised700000,zhang2024selfssl,zhang2024masked,wang2025timedart}. Despite the many differences in public datasets' hardware, protocols, and labeling schemes, IMU signals are still governed by the same wrist kinematics, which suggests that representations learned from broad global motion may transfer to fine-grained personalized recognition.

In this paper, we present \textbf{TransfHAR}, a self-supervised wrist IMU framework for on-demand activity recognition. TransfHAR enables personalized activity recognition from a smartwatch, customized to users' daily mannerisms and preferences. By pooling global motion data from public datasets, we show that training a ViT-1D encoder with self-supervised learning can learn general IMU structure that links different levels of activity types and granularities, in which simple linear probes can adapt effectively to new downstream vocabularies. Our central insight is that global coarse wrist motion provides an effective foundation for fine-grained activities that never appear during pretraining. Although public wrist datasets are dominated by locomotive and postural movements, they still contain recurring motion structure, oscillations, impacts, rotations, and cross-axis coordination patterns that reappear in everyday manipulations, gestures, and task steps.

We evaluate TransfHAR in both offline and interactive settings. Offline, we test transfer ability on three held-out wrist IMU benchmarks spanning manipulative, procedural, and gesture-like activity regimes that are excluded from pretraining, and show that a frozen encoder with a single-layer linear probe trained on each dataset's training split outperforms the supervised baseline on all three datasets, by 8.1 points on SAMoSA, 5.5 on PrISM latte-making, 6.2 points on average across all eight PrISM procedures, and 5.0 on UTD-MHAD, averaging a 6.2-point gain in balanced accuracy. Online, we deploy TransfHAR as a real-time smartwatch system and evaluate it with 10 participants performing seven user-defined activities each, reaching 86.7\% balanced accuracy with only five labeled windows per class and 90.4\% balanced accuracy from a single one-minute recording per class. We show that self-supervised pretraining on broad, coarse wrist motion can serve as a practical foundation for rapid, user-defined, fine-grained activity recognition in interactive systems.

In this paper, we make the following contributions:
\begin{enumerate}
    \item \textbf{TransfHAR}, a self-supervised wrist IMU framework that learns reusable motion representations from heterogeneous public datasets and supports rapid on-device adaptation for user-defined activity recognition.
    \item A real-time smartwatch system that enables users to define activity labels, provide a small number of demonstrations, train a lightweight personalized classifier on-device, and receive real-time, on-demand activity predictions, reaching 86.7\% balanced accuracy using 12.8 seconds of demonstration per class and 90.4\% using one minute per class.
    \item An empirical cross-dataset evaluation showing that frozen self-supervised wrist representations transfer across held-out manipulative, procedural, and gesture-like benchmarks, exceeding supervised baselines by 6.2 balanced-accuracy points on average.
\end{enumerate}

\section{Related Work}

\begin{table*}[t]
\centering
\small
\caption{Public wrist IMU datasets consolidated in TransfHAR and their role in pretraining and probing. All sources are converted to a wrist-only 50\,Hz continuous stream in a shared forward-left-up (FLU) coordinate convention.}
\begin{tabular*}{\textwidth}{@{\extracolsep{\fill}} c c c l c c c c}
\toprule
\textbf{Dataset} & \textbf{Subj.} & \textbf{Hours} & \textbf{Original Placement} & \textbf{Hz} & \textbf{Axes} & \textbf{Classes} & \textbf{Role} \\
\midrule
RecoFit~\cite{Morris2014RecoFit} & 114 & 79.8 & Right forearm & 50 & 6 (A+G) & 28 & Pretrain \\
PAMAP2~\cite{Reiss2012PAMAP2} & 9 & $\sim$10 & Wrist, chest, ankle & 100 & 9 (A+G+M) & 18 & Pretrain \\
Opportunity++~\cite{Ciliberto2021Opportunity} & 4 & 19.75 & Full body (7 IMU + 12 Acc) & 30 & 17 (A+G+M) & 17 & Pretrain \\
WISDM~\cite{Heydarian2023rWISDM} & 51 & 45.9 & Watch (dom.\ wrist) + phone & 20 & 6 (A+G) & 18 & Pretrain \\
Shoaib~\cite{Shoaib2016Complex} & 10 & 6.5 & Wrist (phone) + pocket & 50 & 6 (A+G) & 13 & Pretrain \\
WEAR~\cite{Bock2023WEAR} & 22 & 19 & Both wrists + ankles & 50 & 3 (A) & 18 & Pretrain \\
Capture-24~\cite{Chan2024CAPTURE24} & 151 & 2{,}562 & Dominant wrist & 100 & 3 (A) & 206 & Pretrain \\
UT-Watch~\cite{Bhattacharya2022UTWatch} & 20 & $\sim$17 & Dominant wrist & 50 & 6 (A+G) & 23 & Pretrain \\
\midrule
SAMoSA~\cite{Mollyn2022SAMOSA} & 20 & 14.2 & Dominant wrist & 50 & 9 (A+G+O) & 27 & Probe \\
PrISM-Tracker~\cite{Arakawa2023PrISMTracker} & 14 & $\sim$2.7 & Right wrist & 50 & 9 (A+G+O) & 8--19 & Probe \\
UTD-MHAD~\cite{Chen2015UTDMHAD} & 8 & $\sim$1.6 & Right wrist (1--21), right thigh (22--27) & 50 & 6 (A+G) & 27 & Probe \\
\bottomrule
\end{tabular*}
\label{tab:datasets}
\end{table*}

\subsection{Self-Supervised Pretraining for Wearable Sensing}
Self-supervised learning has become a standard approach for learning transferable representations from unlabeled data across language~\cite{devlin2019bert,liu2020roberta}, vision~\cite{he2020moco,caron2021dino}, and time series~\cite{yue2022ts2vec,zhang2022selfsupervised} through objectives such as masked prediction~\cite{he2021maskedautoencodersscalablevision} or contrastive learning~\cite{SimCLR}. Wearable sensing is particularly well suited for this regime since large-scale inertial datasets are expensive to annotate, often heterogeneous in sensor configuration and collection protocol, and typically require substantial preprocessing before downstream use.

Prior work has explored a range of self-supervised objectives for wearable signals, including multi-task transformation prediction~\cite{Saeed2019MultiTaskSSL}, cross-dimensional motion prediction~\cite{Taghanaki2020ForecastingHAR}, contrastive learning~\cite{Tang2020ContrastiveHAR,Haresamudram2020CPC,Jain2022ColloSSL}, and masked reconstruction~\cite{Haresamudram2020MaskedReconstruction}. More recent efforts have scaled this direction through broader empirical assessment, larger pretraining corpora, and stronger pipelines~\cite{AssessingSSLHAR,Hang2024SelfSupervised700000,SelfPAB,CrossSensorMaskedHAR,Logacjov2024SSLSurvey}, demonstrating that inertial signals contain substantial latent structure learnable without semantic labels. However, most evaluations still focus on aligned benchmark settings where pretraining and downstream tasks remain similar in activity regime, dataset structure, or label space~\cite{Hang2024SelfSupervised700000,SelfHAR,CrossHAR}, where the downstream model is fine-tuned end to end rather than used as a fixed feature extractor~\cite{Hang2024SelfSupervised700000,SelfHAR}. This leaves open the question central to our work: whether broad self-supervised pretraining on predominantly coarse wrist motion can produce a frozen representation that remains useful for qualitatively different fine-grained manipulative, gestural, procedural, and user-defined activities.

\subsection{Limited-Label Adaptation in HAR}
Most HAR systems train supervised task-specific models for a fixed dataset, sensor configuration, and activity vocabulary \cite{Ordonez2016DeepConvLSTM, Hammerla2016DeepHAR} defined a priori. When models do address transfer learning, the focus is on cross-dataset or cross-user generalization, usually within the same activity granularity with relatively small label spaces of fewer than 10 classes. For example, LIMU-BERT adapts BERT-style masked prediction to unlabeled IMU sequences \cite{LIMUBERT}, UniHAR applies physics-informed augmentation grounded in the IMU sensing process \cite{UniHAR}, and CrossHAR combines sensor augmentation with hierarchical self-supervised pretraining to learn more generalizable representations \cite{CrossHAR}. These methods focus on improving robustness to domain shift and label efficiency, but the downstream task structure is largely preserved.
A separate line of work studies label alignment more directly through few-shot learning and invariant feature methods using limited labeled target data \cite{HARDomainAdaptationSurvey, InvariantFeatureLearningHAR, DiggingDeeper, FewShotTransferLearningHAR}. However, these approaches generally assume that target activities share the granularity and environment of the source data, leaving underexplored the case most relevant to interactive wearable systems, where target activities may be sparse, personalized, defined after deployment, and qualitatively different from the coarse motions seen during pretraining.

\subsection{Customization in Wearable Activity Systems}
A complementary thread of work in wearable computing asks how end users can define, customize, or extend recognition systems with minimal effort. Prior smartwatch and on-body systems have shown that wrist-worn sensing can support fine-grained recognition of predefined behaviors, including finger motions on commodity smartwatches \cite{Serendipity}, bio-acoustic hand and object interactions \cite{ViBand}, and daily hand activities such as writing and stirring \cite{LaputFineGrainedHandActivity,Mollyn2022SAMOSA}. This literature establishes the feasibility of rich wrist-level interaction, but typically assumes that the activity set is specified during system design and training.

In user customization work, there has been a focus on using demonstration-based learning, template- and distance-based recognition, and lightweight personalized models. uWave, for example, supports user-defined gestures from a few demonstrations using template-based matching \cite{uWave}. More recent smartwatch work brings this idea closer to on-device adaptation. Xu et al. show that users can extend an existing wrist-worn recognizer with one to five examples by attaching a lightweight user-specific branch to a supervised pretrained model while preserving the original gesture set \cite{xu2022enabling}. Building on these approaches to user customization, TransfHAR supports open, user-defined activity vocabularies drawn from a general self-supervised motion prior, extending beyond predefined gesture sets and single interaction domains.

\section{Datasets}
\begin{table}[t]
\centering
\small
\caption{Motor complexity taxonomy by signal structure.}
\setlength{\tabcolsep}{6pt} 
\begin{tabularx}{\columnwidth}{c l >{\raggedright\arraybackslash}X >{\raggedright\arraybackslash}X}
\toprule
\textbf{Level} & \textbf{Category} & \textbf{Signal Character} & \textbf{Examples} \\
\midrule
L1 & Quasi-static & Gravity-dominated, slow postural drift, low energy above DC & Sitting, Standing, Lying, Watching TV, Car driving \\
L2 & Locomotor & Quasi-periodic, dominant spectral peak & Walking, Running, Cycling, Repetitive exercise \\
L3 & Manipulative & Aperiodic, variable amplitude, no dominant frequency & Chopping, Brushing teeth, Typing, Drilling \\
L4 & Procedural & L3-like per window, separable only with temporal context & Action steps for cooking, Latte-making \\
\bottomrule
\end{tabularx}
\label{tab:taxonomy}
\end{table}

\subsection{Normalization and Alignment}
TransfHAR uses pooled public wrist IMU data, which differ substantially in sensor hardware, sampling rate, placement, channel availability, axis convention, and units. We therefore standardize all sources in Table~\ref{tab:datasets} into a common smartwatch-oriented representation. Each dataset retains a wrist-only IMU stream, resampled at 50 Hz with a shared forward-left-up (FLU) axis convention and consistent physical units. We use the dominant wrist as the canonical reference when available, since it is the most common placement across our sources and best matches smartwatch deployment. In total, the consolidated corpus spans 381 subjects recorded in homes, laboratories, free-living settings, and workshops. All dataset annotations and benchmarks are made publicly available at: \url{https://github.com/Abradshaw1/IMU_LM_Data}.

\subsection{Activity Taxonomy}
Beyond sensor alignment, a second challenge in pooling wrist datasets is inconsistency in how activities are described. HAR papers frequently label datasets as coarse or fine-grained, but these distinctions are often based on class count or naming conventions rather than discriminability from a wrist IMU signal. We therefore define a four-level motor complexity taxonomy grounded in dominant wrist signal characteristics, summarized in Table~\ref{tab:taxonomy}. 

\begin{figure}[t]
  \centering
  \includegraphics[width=\columnwidth]{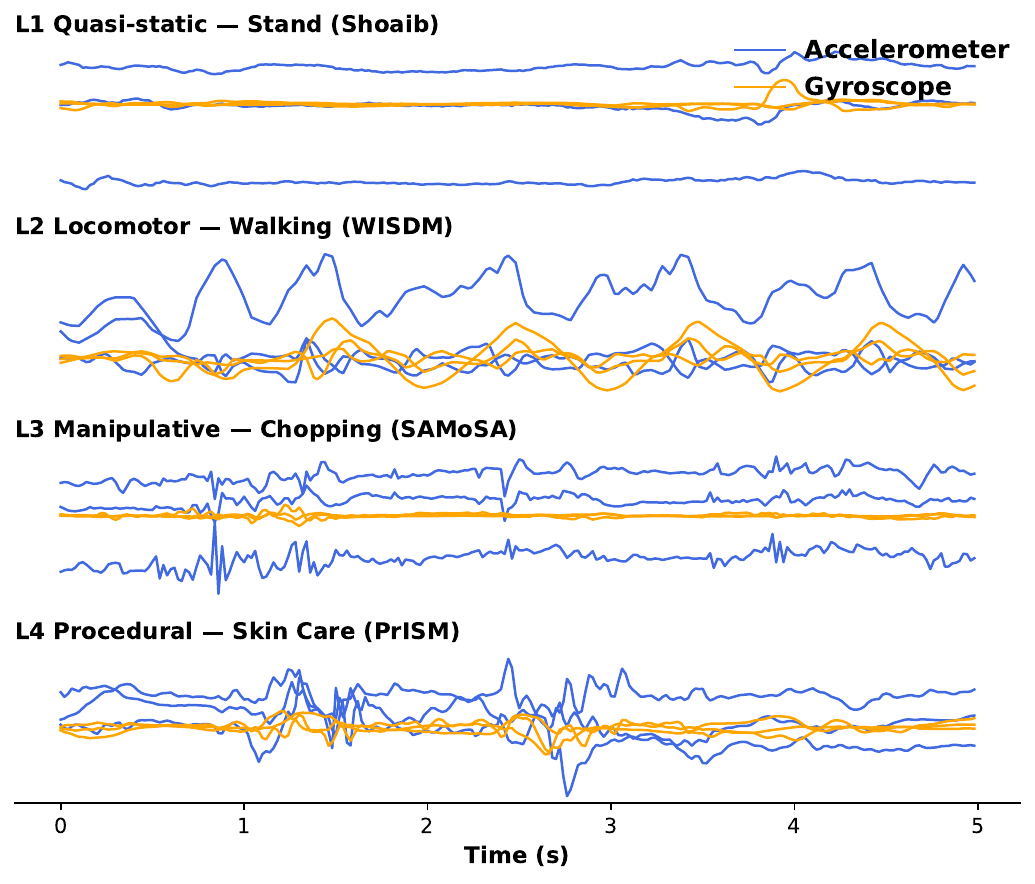}
  \caption{Representative 5-second wrist IMU windows for each motor complexity level from public datasets.}
  \Description{Four stacked time-series panels covering zero to five seconds, one per motor complexity level, each overlaying blue accelerometer channels and orange gyroscope channels. The L1 stationary example of standing from Shoaib is almost flat with separated constant offsets. The L2 locomotor example of walking from WISDM shows smooth repeating oscillations at a steady rate. The L3 manipulative example of chopping from SAMoSA shows irregular spikes of varying height with no repeating cycle. The L4 procedural example of skin care from PrISM shows sparse large excursions separated by quiet stretches.}
  \label{fig:taxonomy_signals}
\end{figure}

\begin{figure}[t]
  \centering
  \includegraphics[width=\columnwidth]{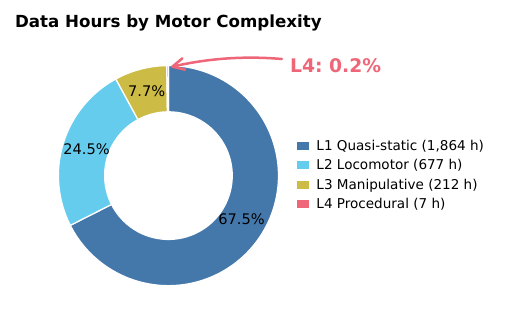}
  \caption{Distribution of data hours by motor-complexity level. The data landscape is dominated by L1/L2 activities, with comparatively limited L3 and negligible L4 coverage.}
  \Description{A donut chart of pretraining hours by motor complexity level. L1 quasi-static accounts for 1,864 hours or 67.5 percent, L2 locomotor for 677 hours or 24.5 percent, L3 manipulative for 212 hours or 7.7 percent, and L4 procedural for only 7 hours or 0.2 percent, with the L4 sliver too thin to see without the annotated callout pointing to it.}
  \label{fig:hours_by_level}
\end{figure}

L1 activities are gravity-dominated with minimal energy above DC and are often separable with simple statistics. L2 activities exhibit quasi-periodic structure with a dominant spectral peak, and include repetitive exercises whose discriminative signature is periodic regardless of semantic label. L3 activities are aperiodic with variable-amplitude bursts, no stable dominant frequency and typically require learned representations. L4 activities are locally similar to L3 within short windows but become separable only with temporal context across an ordered sequence of steps. Figure~\ref{fig:taxonomy_signals} visualizes representative windows from each level, showing that the defining distinctions are signal-level rather than semantic. Figure~\ref{fig:hours_by_level} shows the distribution of L types after aggregating all the datasets, exposing a practical imbalance in public wrist IMU data. Most large datasets emphasize posture, locomotion, and exercise at L1/L2, while fine-grained manipulative and procedural motions at L3/L4 remain much scarcer.


\section{TransfHAR Framework}

\begin{figure*}[t]
    \centering
    \includegraphics[width=1\linewidth]{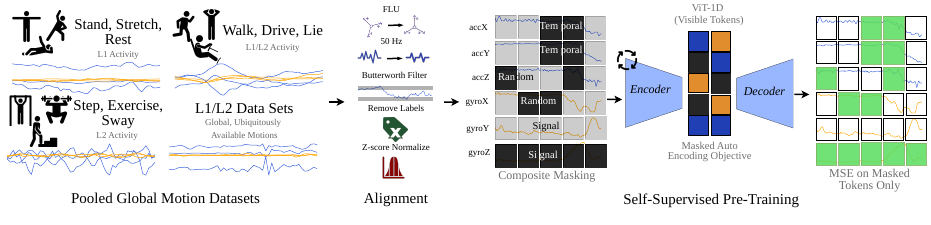}
    \caption{Stage A, self-supervised pretraining on global wrist motion. Public wrist IMU datasets, dominated by coarse L1/L2 activities, are pooled and aligned into a shared representation through a common forward-left-up axis convention, resampling to 50\,Hz, low-pass filtering, label removal, and per-channel z-score normalization. The aligned accelerometer and gyroscope channels are tokenized independently, after which a composite masking strategy hides 65\% of tokens under random, temporal, or whole-channel patterns. A ViT-1D encoder processes only the visible tokens and a lightweight decoder reconstructs the sequence, with the mean squared error loss computed over masked positions alone.}
    \Description{A four-stage pipeline read left to right. The first stage, labeled pooled global motion datasets, shows four panels pairing activity pictograms with overlaid blue and orange sensor traces: an L1 panel covering standing, stretching, and resting, a mixed L1 and L2 panel covering walking, driving, and laying, an L2 panel covering stepping, exercise, and swaying, plus a grayed panel standing for the remaining globally available L1 and L2 datasets. The second stage, labeled alignment, is a vertical block listing a forward-left-up axis conversion, resampling to 50 hertz, Butterworth filtering, label removal, and z-score normalization. The third stage shows the six accelerometer and gyroscope channels stacked as rows of patch tokens with blacked-out regions marked temporal, random, and signal to indicate the three composite masking modes. The masked sequence passes into a trapezoidal ViT-1D encoder operating on visible tokens only, then through a column of shuffled token squares into a smaller decoder, and the final stage shows the reconstructed token grid with masked positions highlighted in green to indicate that mean squared error is computed on those positions only.}
    \label{fig:pretraining}
\end{figure*}

We design TransfHAR as a two-stage framework. In Stage A (Figure~\ref{fig:pretraining}), we pretrain a ViT-1D encoder on pooled public wrist IMU data (3-axis accelerometer and 6-axis accelerometer + gyroscope variants) using a self-supervised masked reconstruction objective over L1/L2 activities, where public data is most abundant. In Stage B (Figure~\ref{fig:deployment}), we freeze this encoder and train a lightweight linear probe on top of its representations to classify user-defined L3/L4 activities. We port the frozen encoder from Stage A and an adaptable Stage B into a single framework on a smartwatch, where users can define activities, record a few demonstrations, train the probe, and receive real-time predictions over their personalized label set. In the remainder of this section, we describe the pretraining framework, the downstream adaptation mechanism, and the watch-based personalization workflow.

\subsection{Self-Supervised Pretraining on Global Motion}
\subsubsection{Pretraining motivation}
Public wrist-IMU datasets overrepresent coarse whole-body motion because locomotion, exercise, and daily-living activities are easy to collect at scale. These data come from the same wrist-worn modality as the downstream tasks we target, so the pretraining distribution is shifted mainly in label granularity rather than in hardware or body placement. Fine-grained downstream activities such as manipulations, gestures, and procedural steps are rarely labeled in public corpora, yet they are composed of the same lower-level wrist motion primitives that appear throughout coarser recordings under different or absent category names. A self-supervised objective exploits this overlap by discarding labels and reconstructing masked signal patches~\cite{he2021maskedautoencodersscalablevision, xu2025lsm2}, learning temporal structure and cross-axis coordination shared across activity types.

\subsubsection{Encoder architecture}
We adopt a 1D Vision Transformer (ViT-1D) adapted from LSM-2~\cite{xu2025lsm2} for wrist IMU signals. The model operates directly on raw time-series inputs using a channel-independent patch tokenization scheme. A shared 1D convolutional kernel (kernel size and stride of 4 samples) is applied independently to each sensor channel, producing 32 patch tokens per channel that are concatenated into a single token sequence. To preserve both temporal and sensor-axis structure, we use a learned 2D positional encoding that decomposes into separate embeddings for time and channel identity. The combined positional embedding is added to each token, allowing the transformer to distinguish temporal position within a channel and the originating sensor axis. The token sequence is processed by a pre-norm transformer encoder with 12 layers, 384-dimensional hidden states, and 6 attention heads, with a 4× expansion in the feed-forward layers and GELU activations. Following the masked autoencoder design, masked tokens are removed prior to the encoder during pretraining so that the transformer operates only on visible tokens. During inference, all tokens are processed and mean-pooled to produce a single 384-dimensional embedding per window.

\subsubsection{Input representation and windowing}
The continuous IMU signal is segmented into fixed-length windows of 2.56 seconds (128 samples at 50 Hz) with 50\% overlap; we validate this choice with a window-length ablation in Appendix~\ref{app:window}. Each window is normalized per channel using z-score normalization. We apply a 4th-order Butterworth low-pass filter at 24 Hz prior to resampling to reduce aliasing and high-frequency noise from resampling and hardware variability. We consider two input configurations, a 3-axis variant uses tri-axial accelerometer signals, and a 6-axis variant that additionally includes tri-axial gyroscope measurements. In both cases, we use channel-independent patching to produce 32 tokens per channel, resulting in 96 tokens for the 3-axis setting and 192 tokens for the 6-axis setting, all with embeddings having dimension 384. The 3-axis encoder is pretrained on a substantially larger corpus than the 6-axis encoder, since the largest public wrist sources (e.g.,\ Capture-24) are accelerometer-only. Retaining both variants lets us study the downstream value of gyroscope signals alongside the effect of pretraining corpus size. Only corresponding L1/L2 activities were selected from each pretrain dataset (Table~\ref{tab:datasets}), with all activity labels discarded during pretraining.

\subsubsection{Training details}
\begin{figure*}[t]
    \centering
    \includegraphics[width=\linewidth]{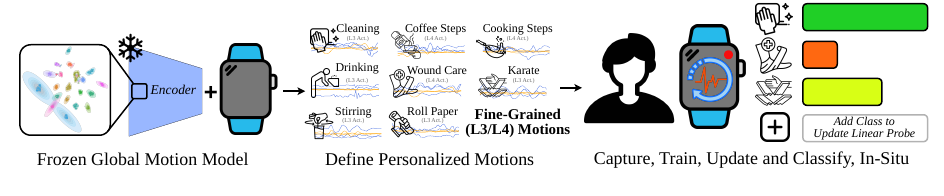}
    \caption{Stage B, on-demand task learning on the watch. The encoder pretrained in Figure~\ref{fig:pretraining} is frozen and paired with the smartwatch (left). Users define their own fine-grained L3/L4 activities, spanning manipulations such as stirring and procedural sequences such as coffee-making steps, and record a small number of in-situ demonstrations for each (center). The watch then streams IMU data, encodes each window with the frozen encoder, and classifies it with the on-device linear probe to return real-time confidence over the personalized vocabulary, while an add-class control expands the label set and retrains the probe without modifying the encoder (right).}
    \Description{A three-stage deployment diagram read left to right. The first stage, labeled frozen global motion model, shows a dense multicolored embedding scatter magnified from a trapezoidal encoder block marked with a snowflake to indicate freezing, joined by a plus sign to a smartwatch. The second stage, labeled define personalized motions, is a grid of activity cards each pairing a line drawing with a short blue and orange sensor trace and a tag giving its complexity level, covering cleaning, coffee steps, cooking steps, drinking, wound care, karate, stirring, and rolling paper, with a final cell naming the whole group fine-grained L3 and L4 motions. The third stage, labeled capture, train, update and classify in situ, shows a person wearing the watch as it records a live signal, beside a ranked column of predicted activities drawn as activity icons with colored confidence bars, and beneath them a plus button labeled add class to update linear probe.}
    \label{fig:deployment}
\end{figure*}

We pretrain the encoder using a masked autoencoder (MAE) objective~\cite{he2021maskedautoencodersscalablevision}. At each training step, 65\% of patch tokens are selected for masking and removed from the encoder input, so the transformer processes only the remaining visible tokens. A lightweight decoder (4 layers, 192-dimensional hidden size, 4 attention heads) reconstructs the masked patches after reinserting learnable mask tokens and restoring positional structure. The training loss is mean squared error computed only over masked patch positions, and the decoder is discarded after pretraining. To improve robustness to temporal and channel structure, we use a composite masking strategy that randomly selects one of three modes per sample: (i) random masking over all tokens, (ii) temporal masking across all channels at selected time steps, and (iii) signal masking across all time steps for selected channels. Each mode masks the same fraction of tokens (65\%), keeping the visible token count constant. In preliminary experiments, we observed that combining random, temporal, and channel-level masking produced more stable downstream transfer than random token masking alone. This design is also motivated by prior findings that structured masking improves representation quality for IMU data, as explored in LSM-2~\cite{xu2025lsm2}. 

The 3-axis encoder is pretrained on all datasets listed in Table~\ref{tab:datasets}, while the 6-axis encoder is trained only on datasets that include gyroscope signals. Datasets used for downstream evaluation are excluded entirely from pretraining. We optimize using AdamW~\cite{Loshchilov2019AdamW} with a learning rate of $1.5\times10^{-4}$, $\beta=(0.9, 0.95)$, and weight decay $10^{-4}$, with a cosine learning rate schedule and 1{,}000-step warmup. Training uses mixed precision (FP16), batch size 512, and gradient clipping at norm 1.0 on 4 NVIDIA H100 GPUs. We validate on a 5\% subject-held-out split and apply early stopping based on validation reconstruction loss. The encoder checkpoint with the lowest validation loss is used for all downstream experiments.

\subsection{On-Demand Task Learning for User-Defined Activities}
\subsubsection{Participant-specific linear probe and watch workflow}
Figure~\ref{fig:deployment} illustrates the transfer and personalization stage (Stage B). After Stage A, the encoder is frozen and a lightweight linear probe is attached on top for transfer learning. Both the 3-axis and 6-axis encoders produce a 384-dimensional mean-pooled embedding per window, which serves as input to the probe. The probe is a single fully connected layer of size $384 \times C + C$, where $C$ is the number of current user-defined activity classes. During adaptation, only the probe parameters are updated while the encoder remains fixed.

On the watch, users define an activity label, record examples in situ, and trigger probe training. During recording, the watch buffers IMU data at 50\,Hz, segments it into 2.56-second windows with 50\% overlap, and encodes each window with the frozen encoder to produce embedding-label pairs for the current activity set. During live use, incoming windows are encoded and classified by the current probe in real time, and users can iteratively add examples or expand their activity vocabulary without retraining the encoder.

\subsubsection{Probe adaptation and training}
Because the probe is a single linear layer, retraining is fast and can be repeated whenever the user adds examples or changes the activity vocabulary. Probe training uses AdamW with learning rate $5\times10^{-4}$, weight decay $10^{-4}$, gradient clipping at norm 1.0, mixed precision (FP16), and batch size 256. We use inverse-frequency class-weighted cross-entropy loss, select checkpoints by validation balanced accuracy, and apply early stopping with patience 40.

\subsection{Implementation and Deployment}
\subsubsection{Prototype pipeline.}
We implement TransfHAR on Apple Watch Series 10 (Figure~\ref{fig:deployment}). The application is written in Swift and uses Core Motion for IMU acquisition, local buffering, windowing, on-device encoder inference, storage of embedding-label pairs, probe training, and real-time classification. The ViT-1D encoder is trained offline in PyTorch and converted to Core ML for deployment, allowing the full interaction loop of capture, train, and classify to run locally on the watch without requiring a paired phone or remote server.

\subsubsection{Latency and footprint.}
The system is designed so that expensive representation learning is performed offline, while only lightweight personalization occurs on-device.
Predictions are generated from 2.56-second windows with 50\% overlap, and once a window is available, encoder inference and probe prediction complete in approximately 23.8~ms on Apple Watch Series 10, yielding a refresh interval of 1.3~s including stride. For a 7-class set, probe training completes in 709, 706, 794, and 786~ms at $K=1,5,10,20$ respectively, fast enough to run between demonstrations. The deployment footprint remains small because the watch stores only the encoder weights, current probe parameters, label set, and cached embedding-label pairs; the encoder has 21.4M parameters, requires 8.83~GFLOPs per window, and occupies approximately 40.8~MB at FP16, while the probe adds only 385 parameters per additional class. Apple's watchOS does not expose fine-grained thermal or power telemetry, so we report latency and GFLOPs as compute-cost proxies rather than direct battery or thermal measurements.

\section{Offline Evaluation}

\begin{figure*}[t]
  \centering
  \includegraphics[width=\textwidth]{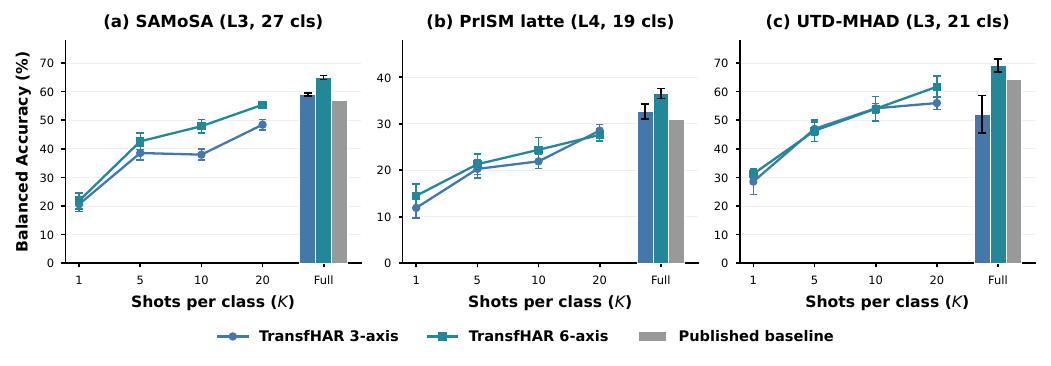}
  \caption{Balanced accuracy as a function of shots per class ($K$ = 1 shot = 2.56s) on three held-out probe targets, with the full-split setting as the bar chart. Supervised baselines are shown as gray bars for each dataset.}
  \Description{Three panels sharing a layout, covering SAMoSA with 27 classes, PrISM latte with 19 classes, and UTD-MHAD with 21 classes. In each panel, balanced accuracy is plotted against one, five, ten, and twenty shots per class as two lines with error bars for the 3-axis and 6-axis encoders, and the full-split condition appears at the right as a group of three bars giving 3-axis, 6-axis, and the supervised baseline. Both curves climb steeply from one to five shots and then rise more slowly, and in all three panels the 6-axis full-split bar stands above the gray baseline bar.}
  \label{fig:fewshot_scaling}
\end{figure*}

In order to  evaluate whether representations pretrained on broad L1/L2 wrist motion transfer to held-out fine-grained L3/L4 activities, we probe the frozen encoder on three public wrist IMU benchmarks spanning manipulative, procedural, and gesture-like activities. 

\subsection{Datasets}
\subsubsection{SAMoSA}
SAMoSA~\cite{Mollyn2022SAMOSA} contains 27 everyday activity classes collected from 20 participants wearing a smartwatch in home and workshop environments. The dataset includes synchronized 50\,Hz wrist IMU and audio; we use only the motion stream. Activities span kitchen, bathroom, and workshop manipulations, placing it primarily in the L3 regime.

\subsubsection{PrISM-Tracker}
PrISM-Tracker~\cite{Arakawa2023PrISMTracker} is a procedural activity dataset collected from smartwatch motion and audio during multi-step daily tasks. The full dataset contains eight procedures, each composed of multiple labeled task steps. We evaluate TransfHAR on PrISM in two ways. First, we use the latte-making procedure, which contains 19 discrete steps performed across 23 sessions from 14 unique participants, for both few-shot and full-split comparison. Second, we compare against the PrISM motion-only baselines across all procedures in the dataset. These tasks are locally similar within short windows and must be distinguished within flexible multi-step sequences, placing PrISM primarily in the L4 regime.

\subsubsection{UTD-MHAD}
UTD-MHAD~\cite{Chen2015UTDMHAD} is a multimodal human action dataset containing depth, RGB, skeleton, and inertial recordings for 27 actions performed by 8 subjects. We use only the inertial modality, restricted to the 21-class wrist subset (UTD1) and compare with the published inertial baseline from Yang et al.~\cite{Yang2022ActivityGraph}. This subset contains short, burst-like gesture and exercise motions such as arm swings, drawing motions, knocks, and throws, representative of the L3 regime.

\subsection{Protocol}
For probe training and evaluation, we use each dataset's evaluation protocol. Since PrISM does not report all-procedure performance, we pull their original code and run their motion-only model using their data and protocol. For SAMoSA we also pull their code and run their model under the same protocol to produce baselines. We evaluate both 3-axis and 6-axis encoders, each paired with its corresponding pretrained checkpoint.

\subsubsection{Few-shot and full-split conditions}
We evaluate two probe training regimes. In the \emph{few-shot} regime, the probe is trained on $K \in \{1,5,10,20\}$ labeled windows per class, sampled uniformly at random from the training split rather than taken contiguously from its start. One shot is one 2.56-second window, so $K=1,5,10,20$ correspond to 2.56, 12.8, 25.6, and 51.2 seconds of labeled data per class. Only the training split is subsampled; validation and test sets are unchanged across all $K$. In the \emph{full-split} regime, the probe is trained on all windows in the dataset's training partition. In both regimes the encoder is frozen and only the probe is optimized, and evaluation follows each published baseline's protocol. All results are reported as the mean over five random seeds with the standard deviation in parentheses.

\subsubsection{Metrics}
We report balanced accuracy to account for class imbalance across benchmarks, defined as: $\mathrm{Balanced\ Accuracy} = \frac{1}{C}\sum_{c=1}^{C} \frac{\mathrm{TP}_c}{\mathrm{TP}_c + \mathrm{FN}_c}$. Balanced accuracy is used as the primary model-selection metric during probe training. For supervised comparison, we benchmark against the corresponding motion-only baseline for each dataset. The SAMoSA and PrISM baselines use 9-axis input (accelerometer, gyroscope, and magnetometer/orientation), while the UTD-MHAD baseline uses 6-axis accelerometer and gyroscope input; all are fully supervised.

\subsection{Offline Results}
\label{sec:offline}

\begin{figure}[t]
  \centering
  \includegraphics[width=\columnwidth]{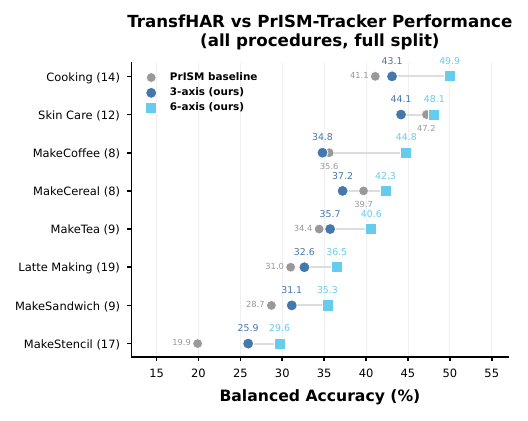}
  \caption{Full-split balanced accuracy across all eight PrISM-Tracker procedures for 3-axis and 6-axis encoders. Number of procedure steps is shown in parentheses.}
  \Description{A horizontal dot plot with one row for each of the eight PrISM-Tracker procedures, ordered from the highest scoring Cooking at the top down to the lowest scoring MakeStencil at the bottom, with the number of steps in each procedure given in parentheses. Every row carries three labeled points, a gray PrISM baseline, a dark blue 3-axis result, and a light blue 6-axis result, plotted against balanced accuracy from 15 to 55 percent. The 6-axis point sits furthest right on every row, the 3-axis point sits between the other two on most rows, and the whole ordering shifts leftward as procedures become harder.}
  \label{fig:prism_all_tasks}
\end{figure}

\subsubsection{Few-shot scaling}
Figure~\ref{fig:fewshot_scaling} shows balanced accuracy as a function of shots per class on the three held-out probe targets, with the full-split setting included as the final point for reference. Across all three datasets, 6-axis performance rises sharply between $K{=}1$ and $K{=}5$, then improves more gradually through $K{=}20$ before reaching its highest value in the full-split setting. For example, on SAMoSA the 6-axis probe increases from 21.8\% ($SD=2.8$) at $K{=}1$ to 42.6\% ($SD=2.9$) at $K{=}5$, and on UTD-MHAD it rises from 31.2\% ($SD=1.7$) to 46.3\% ($SD=3.8$) over the same range. This early jump suggests the frozen encoder already provides class boundaries that additional data refines rather than creates.

As $K$ increases, performance continues to improve but with diminishing returns, reaching 55.4\% ($SD=0.9$) on SAMoSA and 61.7\% ($SD=3.8$) on UTD-MHAD at $K{=}20$, before rising to 64.9\% ($SD=0.7$) and 69.1\% ($SD=2.3$) in the full-split setting. In contrast, gains on PrISM latte-making are smaller, increasing from 14.5\% ($SD=2.5$) to 27.6\% ($SD=1.3$) for 6-axis between $K{=}1$ and $K{=}20$ and reaching 36.5\% ($SD=1.1$) with the full training split. This reflects the underlying task structure. While L3 activities are often separable from short wrist windows, L4 procedural steps are more locally similar and depend more strongly on temporal context beyond a single 2.56-second segment, consistent with our window-length ablation (Appendix~\ref{app:window}), where PrISM latte-making alone benefits from longer 5.12\,s windows. Across datasets, the 6-axis encoder outperforms the 3-axis variant at nearly every supervision level, with the two variants at parity on UTD-MHAD at intermediate shot counts, and the magnitude of the gain depends on the activity type. The gap is larger on SAMoSA, e.g.\ 55.4\% ($SD=0.9$) vs.\ 48.4\% ($SD=1.9$) at $K{=}20$ and 64.9\% ($SD=0.7$) vs.\ 59.0\% ($SD=0.5$) in the full-split setting, where rotational dynamics are important, and smaller on UTD-MHAD at intermediate shot counts, where burst-like motions are already well captured by acceleration alone.

\subsubsection{Full-split baseline comparison}
In the full-split setting (Figure~\ref{fig:fewshot_scaling}), the 6-axis TransfHAR probe exceeds the corresponding supervised baseline on all three datasets: 64.9\% ($SD=0.7$) vs.\ 56.8\% on SAMoSA, 36.5\% ($SD=1.1$) vs.\ 31.0\% on PrISM latte, and 69.1\% ($SD=2.3$) vs.\ 64.1\% on UTD-MHAD. On UTD-MHAD, this full-split advantage is also substantial relative to the 3-axis variant, with 69.1\% ($SD=2.3$) for 6-axis versus 52.1\% ($SD=6.6$) for 3-axis. Across these three benchmarks, the frozen TransfHAR probe exceeds the corresponding published supervised task-specific pipelines. Notably, the gains on SAMoSA and PrISM require only accelerometer and gyroscope inputs against 9-axis baselines, suggesting that self-supervised pretraining recovers useful rotational structure without explicit orientation features. Per-class confusion structure for all three benchmarks is given in Figure~\ref{fig:app_confusion_offline}.

\subsubsection{Transfer across all PrISM procedures}

While our main PrISM results focus on the latte-making benchmark, we additionally evaluate full-split transfer across all eight procedures in the dataset to assess how broadly the learned representation generalizes within the L4 procedural regime. Figure~\ref{fig:prism_all_tasks} shows that the 6-axis encoder matches or exceeds the fully supervised PrISM 9-axis baseline on every procedure, improving balanced accuracy by 6.2 points on average, while the 3-axis encoder is close to parity at 0.9 points and falls below the baseline on three procedures. The largest gains appear on \emph{MakeStencil} (+9.7\,pp) and \emph{MakeCoffee} (+9.2\,pp), suggesting that gyroscope information is especially valuable for procedures whose steps remain difficult to separate from short wrist windows alone.

At the same time, the figure shows that transfer difficulty varies substantially across procedures. Performance is strongest on \emph{Cooking} (49.9\%) and \emph{Skin Care} (48.1\%), but remains much lower on \emph{MakeStencil} (29.6\%) and \emph{MakeSandwich} (35.3\%), where individual steps are likely more locally similar within short windows. This pattern highlights the limitation of window-level classification for L4 activities, where procedural steps are often distinguished more by temporal context than by instantaneous wrist motion.

\subsubsection{Architecture-matched comparison}
\begin{table}[t]
\centering
\small
\caption{Architecture-matched comparison on the full training split, all using the same 6-axis ViT-1D. Balanced accuracy, mean (SD) over five seeds.}
\setlength{\tabcolsep}{4pt}
\begin{tabular}{lccc}
\toprule
\textbf{Training regime} & \textbf{SAMoSA} & \textbf{PrISM latte} & \textbf{UTD-MHAD} \\
\midrule
From scratch (no SSL) & 28.8 (3.4) & 16.2 (3.1) & 46.4 (2.2) \\
Pretrained, fine-tuned & 61.4 (0.8) & 34.4 (0.8) & 65.4 (3.6) \\
Pretrained, frozen + probe & \textbf{64.9 (0.7)} & \textbf{36.5 (1.1)} & \textbf{69.1 (2.3)} \\
\bottomrule
\end{tabular}
\label{tab:matched_comparison}
\end{table}
Published baselines differ from our encoder in both architecture and input modality, so we additionally train two supervised variants of the same 6-axis ViT-1D under the identical full-split protocol (Table~\ref{tab:matched_comparison}). Training from scratch falls far below both pretrained variants despite identical capacity, indicating that transfer stems from the learned representation rather than the architecture. Fine-tuning the encoder offers no advantage over freezing it, suggesting that a linear probe may be better suited for model adaptation.

\section{On-Demand Personalization Study}
The offline evaluations in Section~\ref{sec:offline} use fixed public benchmarks. We now evaluate TransfHAR in the intended setting where users define their own wrist activities on a smartwatch and obtain recognition from a few demonstrations using the frozen encoder and a per-participant linear probe.

\begin{figure}[t]
    \centering
    \includegraphics[width=\linewidth]{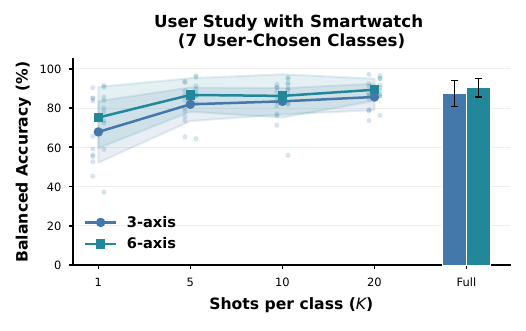}
    \caption{Personalization performance across participants as a function of shots per class ($K$ = 1 shot = 2.56s), with the full-session setting included as the bar chart. Points show per-participant balanced accuracy and lines show participant means for the 3-axis and 6-axis encoders.}
    \Description{A plot of balanced accuracy from zero to one hundred percent against one, five, ten, and twenty shots per class for the seven user-chosen classes, with the full-session condition shown at the right as two bars with error bars. Two mean lines with shaded bands cover the 3-axis and 6-axis encoders, and faint individual participant points scatter around them. Both lines rise steeply between one and five shots and then flatten, the 6-axis line stays above the 3-axis line throughout, and the participant scatter is widest at one shot and tightens as examples are added.}
    \label{fig:userstudy_fewshot}
\end{figure}

\subsection{Study Design}

\subsubsection{Participants and apparatus.}
We recruited 10 participants (7 right-handed, 3 left-handed; mean age 26.1, $SD=2.0$) from our university community. The study was approved by the Institutional Review Board, and participants received \$10 compensation. Each session lasted approximately 30 minutes. Participants were instructed to wear the Apple Watch on their dominant wrist. Accordingly, 7 participants wore the watch on the right wrist and 3 on the left wrist.

\subsubsection{Activity selection and data collection procedure.}
Participants were shown a bank of 55 candidate activities (Appendix Table~\ref{tab:activity_bank}) spanning hand and air gestures (e.g., drawing a heart in the air, waving goodbye), object manipulation (e.g., stirring coffee with a spoon, opening a laptop lid), workspace actions (e.g., typing on a keyboard, writing with a pen), and personal motions (e.g., brushing hair, filing nails). The bank was designed to encourage choosing L3/L4 granularity and discourage selection of L1/L2 coarse activities like walking or jogging. Participants selected 7 activities from this list or defined their own, after being told: \textit{Below is a list of example daily wrist activities to illustrate the level of granularity we are targeting. Please choose activities from this list, or define your own, based on what would be most useful in your daily life.}

For each activity, we recorded 3 separate 1-minute sessions of continuous IMU data while the participant repeatedly performed that activity. Between sessions, participants removed the watch and took a 30-second break before re-wearing it, introducing natural variation in watch placement and fit. During recording, the participant selected the activity label in the interface, pressed \emph{Start}, performed the activity continuously, then stopped the recording.

\subsection{Evaluation Protocol}

\subsubsection{Segmentation and splits.}
All recordings were segmented into 2.56-second windows (128 samples at 50\,Hz) with 50\% overlap, matching the setup used throughout the paper. Evaluation was performed separately for each participant because the class vocabulary differed across users. The study yields 210 one-minute trials (10 participants $\times$ 7 activities $\times$ 3 sessions), each producing approximately 45 windows. Evaluation is cross-session round robin: for each participant and activity, one session trains the probe and the two held-out sessions form the test set, rotating over all three assignments. Reported values are cross-session and post re-wearing.

\subsubsection{Few-shot and full-session probing.}
We evaluate two probing regimes on top of the frozen encoder. In the \emph{full-session} setting, the linear probe is trained using all windows from the designated training session for each of the participant's 7 activities. Since each training session is one minute long, this condition reflects the performance achievable from a single full recording per class. In the \emph{$K$-shot} setting, we subsample $K$ training windows per class from that same session, sampled uniformly at random across the session rather than from its beginning. We evaluate $K \in \{1,5,10,20\}$ with the test set fixed as the two held-out sessions.

\subsection{Results}
\begin{figure}[t]
    \centering
    \includegraphics[width=\linewidth]{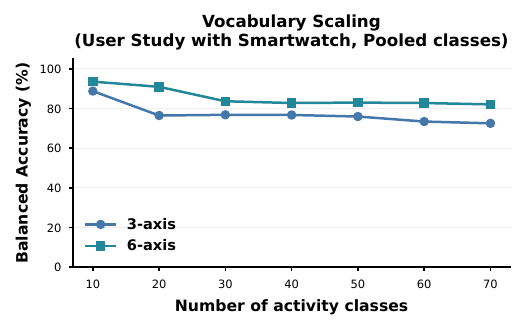}
    \caption{Vocabulary scaling in the user study with all participants' activities pooled into a single non-personalized probe, and balanced accuracy measured as the number of classes the probe must distinguish from 10 to 70. Accuracy declines gradually, with most of the drop occurring below 30 classes.}
    \Description{A plot of balanced accuracy against the number of pooled activity classes at ten, twenty, thirty, forty, fifty, sixty, and seventy classes. The 6-axis line starts near 94 percent and steps down to roughly 82 percent, where it stays almost flat from thirty classes onward. The 3-axis line starts near 89 percent, drops more sharply to the mid seventies by twenty classes, and then declines slowly to roughly 72 percent, leaving a widening gap between the two encoders.}
    \label{fig:userstudy_vocab_scaling}
\end{figure}

We report user-study results on few-shot adaptation, vocabulary scaling, and error structure. Because each participant defined a different set of 7-class activities, the results are aggregated across participants, with participant-specific activity labels provided in the Appendix Table~\ref{tab:userstudy_activity_map}.

\subsubsection{Personalization and few-shot performance}

\begin{figure*}[t]
  \centering
  \includegraphics[width=\textwidth]{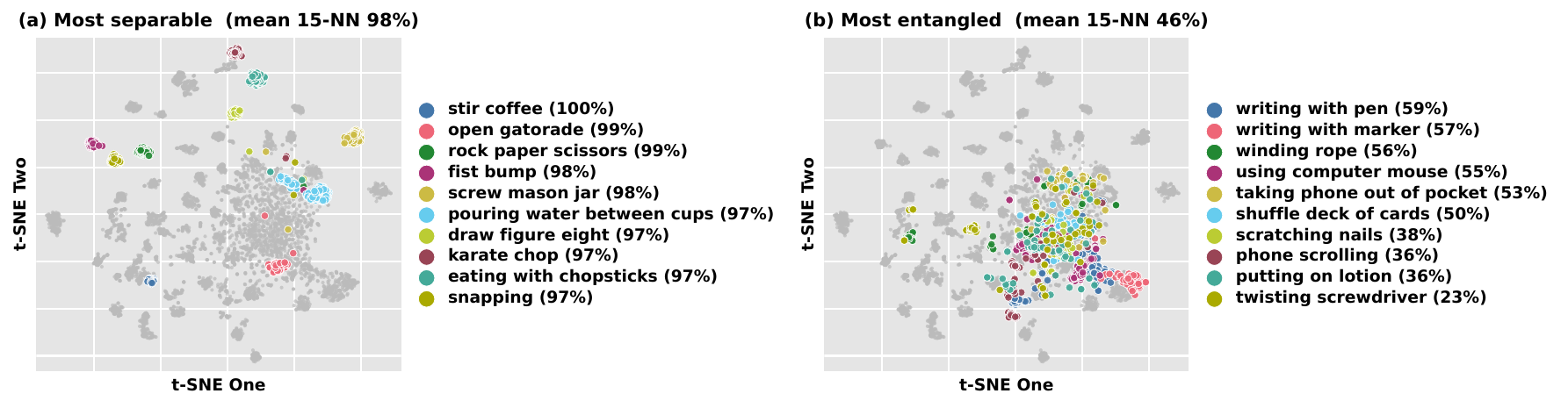}
  \caption{t-SNE of user-study embeddings from the frozen 6-axis encoder, showing the (a) ten most separable and (b) ten most entangled activities against all other windows in gray. Both panels share one projection. Percentages are neighborhood purity, the fraction of each window's 15 nearest neighbors in the 384-dimensional embedding sharing its label.}
  \Description{Two t-SNE scatter panels sharing one projection, with every other window drawn in gray behind the highlighted classes. Panel a shows the ten most separable activities, which form small tight clusters scattered around the edges of the gray mass and reach a mean fifteen-nearest-neighbor purity of 98 percent, led by stirring coffee at 100 percent along with opening a bottle, rock paper scissors, a fist bump, and screwing a jar lid. Panel b shows the ten most entangled activities, which pile into one dense overlapping region near the center at a mean purity of 46 percent, ranging from writing with a pen at 59 percent down to twisting a screwdriver at 23 percent, and including writing with a marker, winding rope, mouse use, scrolling a phone, and applying lotion.}
  \label{fig:tsne}
\end{figure*}

In the full-session setting, where the probe is trained on one 1-minute recording per class, the 6-axis encoder reaches 90.4\% balanced accuracy ($SD=4.8$) and 89.9\% macro-F1 ($SD=5.3$), while the 3-axis encoder reaches 87.3\% balanced accuracy ($SD=6.6$) and 86.6\% macro-F1 ($SD=7.2$). Figure~\ref{fig:userstudy_fewshot} shows that personalization remains effective even with very limited supervision. With one labeled window per class (2.56 seconds of user data), the 6-axis encoder reaches 75.2\% balanced accuracy on average across participants ($SD=15.6$), compared with 67.8\% for 3-axis ($SD=15.7$). At five shots, performance rises sharply to 86.7\% ($SD=8.5$) for 6-axis and 81.9\% ($SD=8.6$) for 3-axis. Gains continue through $K=20$, reaching 89.3\% and 85.7\%, but the largest improvement occurs between one and five shots. By $K=20$, performance is already within 1.1\,pp of the full-session setting for the 6-axis encoder, suggesting that most of the available personalization benefit can be recovered without collecting a full minute of data per class. Splitting by handedness, right-handed participants (n=7) average 88.7 (5.0) and left-handed participants (n=3) 94.2 (2.3), though the left-handed group is underpowered (Figure~\ref{fig:app_confusion_user}). The 6-axis encoder outperforms the 3-axis encoder at every supervision level, with the largest advantage in the lowest-data regime: 7.4\,pp at $K=1$, narrowing to 3.1\,pp in the full-session setting. Participant variability is highest in the one-shot regime and decreases as more examples are added.

\begin{figure}[t]
    \centering
    \includegraphics[width=\linewidth]{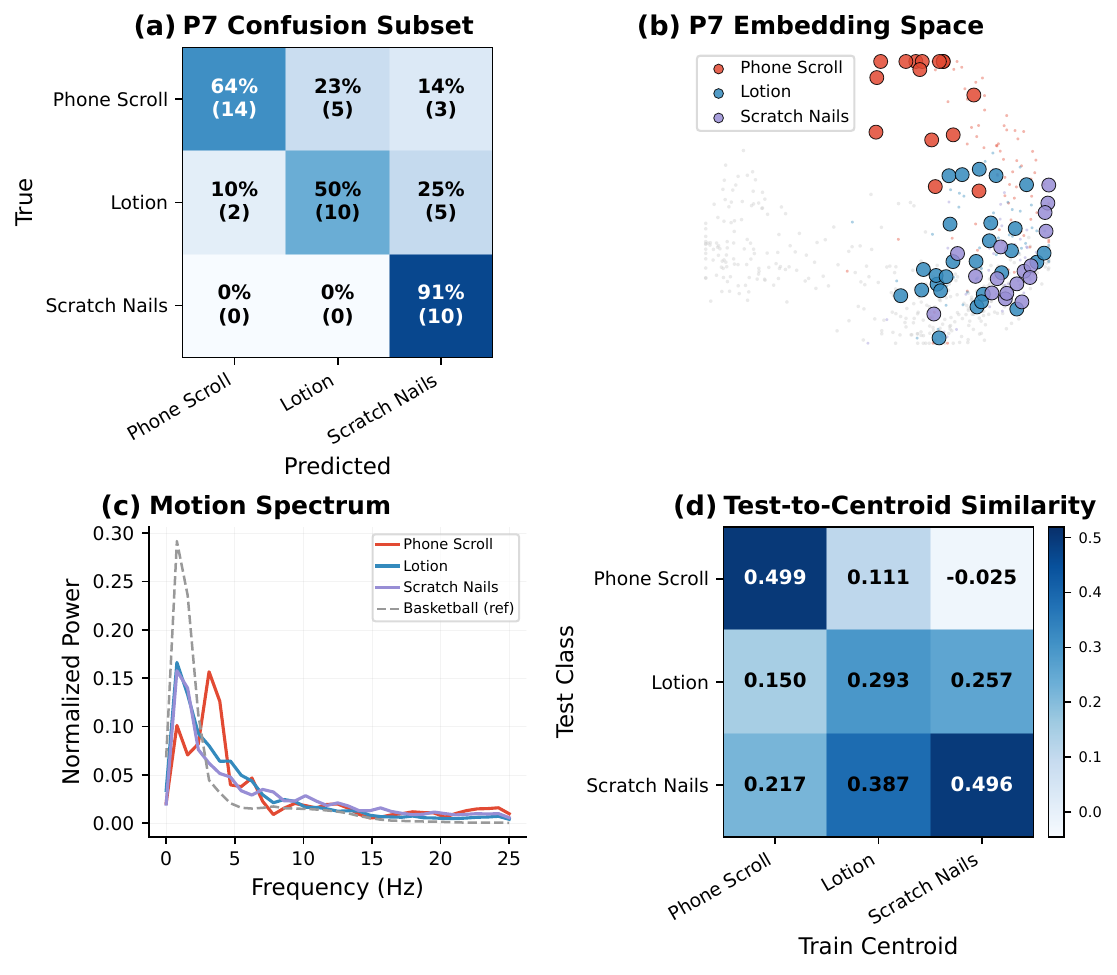}
    \caption{P7 failure analysis in the 6-axis full-session setting. Top left shows the confusion subset for low-performing activities. Top right shows PCA projection of P7 embeddings, with the highlighted classes shown relative to the participant's remaining activities in gray. Bottom left shows normalized acceleration-magnitude spectra. Bottom right shows mean similarity between test samples and train centroids.}
    \Description{Four panels analyzing participant P7. Panel a is a three by three confusion matrix over phone scrolling, lotion, and scratching nails, in which phone scrolling is correct 64 percent of the time, lotion only 50 percent with a quarter of its windows going to scratching nails, and scratching nails 91 percent. Panel b is a PCA scatter in which those three classes occupy one overlapping cloud while the participant's remaining activities sit in gray around them. Panel c plots normalized power against frequency up to 25 hertz for the three confused classes together with basketball shooting as a well-separated reference, and the three confused spectra track one another closely below about five hertz while the reference curve departs sharply. Panel d is a three by three similarity heatmap between test samples and training centroids, where the diagonal values are only weakly higher than the neighboring off-diagonal values, most visibly for lotion.}
    \label{fig:p7_failure}
\end{figure}

\subsubsection{Vocabulary scaling}

Personalization also depends on how many activities the system must distinguish at once. To evaluate this, we pool all participants' data into a single non-personalized model, without participant identity as input, and measure how recognition scales as the class count grows. This pools 70 trials for probing against 140 held-out trials for test, under the same cross-session round robin scheme. Figure~\ref{fig:userstudy_vocab_scaling} shows that balanced accuracy degrades gradually as the class count increases. The 6-axis model drops from 93.6\% at 10 classes to 82.1\% at 70 classes; the 3-axis model drops from 88.8\% to 72.5\%. Most of the decline occurs between 10 and 30 classes, with performance largely plateauing beyond that (the 6-axis model varies by less than 1 point from 30 to 60 classes). Users can therefore expand their label vocabulary over time without retraining the encoder.

\subsubsection{Error structure and failure analysis}
Although overall performance is high, errors are not uniformly distributed. Most participant-activity pairs are recognized near ceiling, while classes below 70\% accuracy cluster within P7, with isolated cases in P1, P4, and P8; the per-class accuracy map and per-participant confusion matrices are given in Appendix~\ref{app:confusion} (Figures~\ref{fig:userstudy_hardmap} and~\ref{fig:app_confusion_user}). Figure~\ref{fig:tsne} ranks every user-defined activity by neighborhood purity in the embedding space. The most separable classes form compact, isolated clusters and are dominated by distinct burst or rotational motions such as stirring, snapping, and screwing a jar lid. The most entangled classes collapse into a shared central region and are predominantly sustained, low-amplitude motions such as writing, scrolling, and applying lotion, whose wrist dynamics differ far less than their semantic labels suggest.

We additionally examine P7 as a representative case (Figure~\ref{fig:p7_failure}). The confusion matrix (a) confirms that errors are concentrated among \textit{Phone Scroll}, \textit{Lotion}, and \textit{Scratch Nails} rather than spread across the full 7-class vocabulary, with \textit{Lotion} most often misclassified as \textit{Scratch Nails} (25\%). The same pattern appears for P4, whose \textit{winding rope} is predicted as \textit{tying a shoelace} 25\% of the time. In PCA space (b), the confused classes occupy a shared local region relative to P7's other activities, indicating overlap within the participant-specific manifold rather than global collapse. Their normalized motion spectra (c) are similar, especially at low frequencies, suggesting that these activities generate closely related wrist dynamics. The test-to-centroid similarity matrix (d) shows weak diagonal margins between confused classes, confirming that errors arise from semantically different activities that produce similar wrist motion which remain difficult to separate from short windows alone.

\section{Discussion}

\begin{figure}[t]
    \centering
    \includegraphics[width=0.95\linewidth]{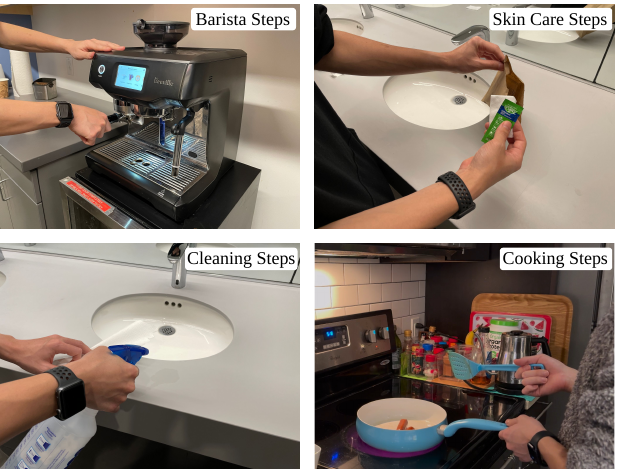}
    \caption{Example on-demand activity scenarios captured with the watch-based system: espresso-making steps, skin care, surface cleaning, and cooking.}
    \Description{A two-by-two grid of photographs, each showing a person wearing a smartwatch while performing a procedural activity, with a label in the corner of each photo. Top left, labeled barista steps, operating the portafilter of an espresso machine. Top right, labeled skin care, holding a skincare packet and tube over a sink. Bottom left, labeled clean steps, spraying a sink with a cleaning bottle. Bottom right, labeled cooking, stirring a pan of food on a stovetop with a spatula.}
    \label{fig:app_scenarios}
\end{figure}

\subsection{Motion Priors from Broad Wrist Data}
Our main finding is that broad wrist motion can act as an anchor for fine-grained recognition. Although Stage A sees only coarse L1 and L2 activities, the resulting encoder transfers to manipulative, gestural, and procedural classes that never appear during pretraining. This suggests that many downstream wrist labels do not require entirely new motion structure to be learned from scratch. Instead, they refine distinctions that are already latent in a representation trained on diverse global motion. This is especially important for wrist IMU, where useful structure must be recovered from temporal dynamics rather than rich spatial context.

\subsection{Application Scenarios for On-Demand Activity Vocabularies}

Our results suggest that users can introduce or revise activity classes at deployment time by providing only a few examples, rather than relying on a fixed, predefined vocabulary. This supports wearable applications where the activity set cannot be fixed in advance. For example, context-aware assistants can adapt to user-specific routines~\cite{DBLP:conf/uist/ArakawaPPLG25}, factory systems can track evolving assembly workflows~\cite{DBLP:conf/petra/SopidisHAAAFB22}, and medical applications can monitor personalized or procedure-specific actions~\cite{DBLP:journals/tmm/BernalYLKMRB18}, including passive motion-based measurement of symptoms such as hyperactivity~\cite{DBLP:journals/imwut/ArakawaAMTSLG23} and tremor~\cite{DBLP:journals/sensors/SigchaPCCGALA21}, without requiring large-scale recollection and retraining. Figure~\ref{fig:app_scenarios} illustrates four such scenarios captured with our system, spanning espresso preparation, skin care, surface cleaning, and cooking.

At the same time, our results highlight a boundary condition for these scenarios. On-demand recognition is most effective for kinematically distinct L3 activities with different burst structure or rotational dynamics, so a barista's pour-over steps separate well because each involves a distinct tool motion, whereas a desk worker's writing and mouse use do not, since both are small sustained hand movements and fall among the most entangled classes in Figure~\ref{fig:tsne}. L4 procedural steps remain harder and only partially solved, since they are locally similar within a single window and separable mainly through temporal context (Appendix~\ref{app:window}). We found these failures to be modality-specific rather than demonstration-specific, with additional examples not helping when two activities produce genuinely overlapping wrist motion. Practical systems should therefore warn users at definition time when a new class falls too close to an existing one in embedding space and offer to merge them, as explored in prior work on representation-driven feedback~\cite{xu2022enabling,DBLP:conf/chi/Wu0BL20,DBLP:journals/imwut/PatidarAGMSVTSSMGA25}.

\subsection{Limitations and Future Work}
TransfHAR has several limitations. First, the current formulation is window-based and therefore cannot model longer temporal structure, which limits performance on procedural L4 tasks whose steps are locally similar but sequentially distinct. A natural next step is to augment the frozen encoder with a lightweight temporal aggregation layer, such as a small recurrent module or state-based model over window embeddings, to capture richer sequential context. Our comparisons also isolate pretraining rather than architecture, so how the frozen representation compares to contemporary self-supervised backbones remains open to future work. Second, our personalization study is short-horizon and controlled. Although we include watch removal and re-wearing, we do not evaluate longer-term behavioral drift or repeated re-personalization over time. Future work should therefore study longitudinal use in more natural settings, including how users revise activity vocabularies and when re-adaptation is needed. Third, while the watch workflow is lightweight, broader adoption questions remain around hardware variation, battery and thermal cost, and interface support to collect better examples. The transformer encoder is also heavier than edge-optimized CNN or LSTM backbones, so model compression and edge-computing optimization remain open for future work. Fourth, our pretraining corpus pools public datasets whose demographic metadata are sparse and inconsistent, so we cannot characterize how well it represents the broader population, and whether the learned representation transfers equally across body types, ages, and motor abilities remains untested. 

\section{Conclusion}
We presented TransfHAR, a framework for on-demand wrist activity recognition built on self-supervised pretraining. By learning broad motion priors from heterogeneous public wrist IMU data and adapting them with a lightweight linear probe, TransfHAR supports rapid personalization to new fine-grained activities without retraining a full model. Our offline experiments across three held-out benchmarks showed that the frozen representation transfers strongly to manipulative, gestural, and procedural tasks, often matching or exceeding supervised baselines. Our user study with a smartwatch further showed that users can define their own activity sets and obtain strong recognition from only a few demonstrations.

\begin{acks}
We thank VESSL AI for providing compute credits that supported this work.
\end{acks}

\bibliographystyle{ACM-Reference-Format}
\bibliography{references}

\appendix

\section{Window Length Ablation}
\label{app:window}

Table~\ref{tab:window_ablation} compares three window lengths under the full-split protocol. The 2.56\,s window used throughout the paper is best on both L3 benchmarks, while the L4 procedural task benefits from the longer 5.12\,s window.

\begin{table}[ht]
\centering
\small
\caption{Window length ablation, frozen 6-axis encoder with a linear probe, full split. Balanced accuracy, mean (SD) over five seeds.}
\begin{tabular}{lccc}
\toprule
\textbf{Dataset} & \textbf{1.28\,s} & \textbf{2.56\,s} & \textbf{5.12\,s} \\
\midrule
SAMoSA (L3) & 50.8 (0.6) & \textbf{64.9 (0.7)} & 61.4 (1.4) \\
UTD-MHAD (L3) & 46.7 (0.3) & \textbf{69.1 (2.3)} & 57.1 (3.5) \\
PrISM latte (L4) & 29.2 (0.3) & 36.5 (1.1) & \textbf{38.1 (2.9)} \\
\bottomrule
\end{tabular}
\label{tab:window_ablation}
\end{table}

\section{Confusion Matrices and Error Maps}
\label{app:confusion}
\begin{figure}[ht]
    \centering
    \includegraphics[width=0.8\linewidth]{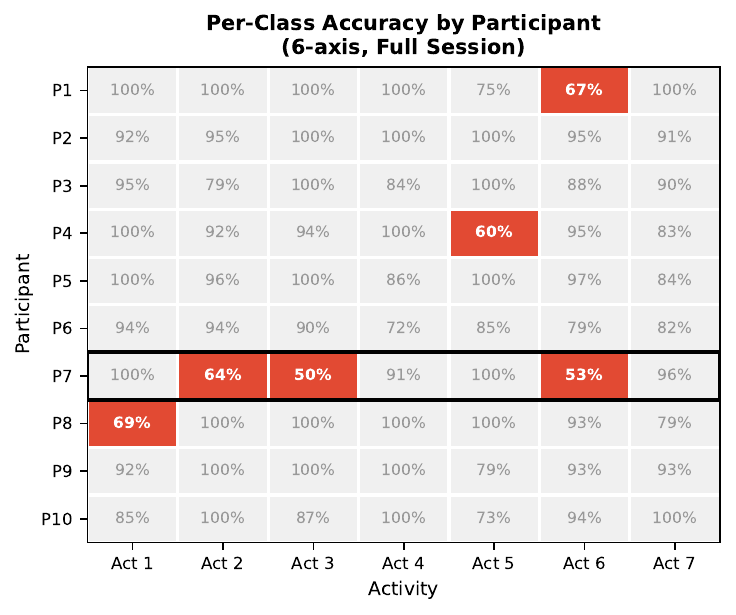}
    \caption{Per-class accuracy for each participant-activity pair, 6-axis full-session setting, with classes below 70\% highlighted.}
    \Description{A ten-by-seven grid with participants P1 through P10 as rows and activity slots one through seven as columns, each cell labeled with per-class accuracy. Most cells read 100 percent or close to it and are shaded pale, while six cells falling below seventy percent are filled red. Three of those red cells lie in the row for P7, which is outlined for emphasis, and the remaining three are isolated cells in the rows for P1, P4, and P8.}
    \label{fig:userstudy_hardmap}
\end{figure}

\begin{figure*}[ht]
  \centering
  \includegraphics[width=\textwidth]{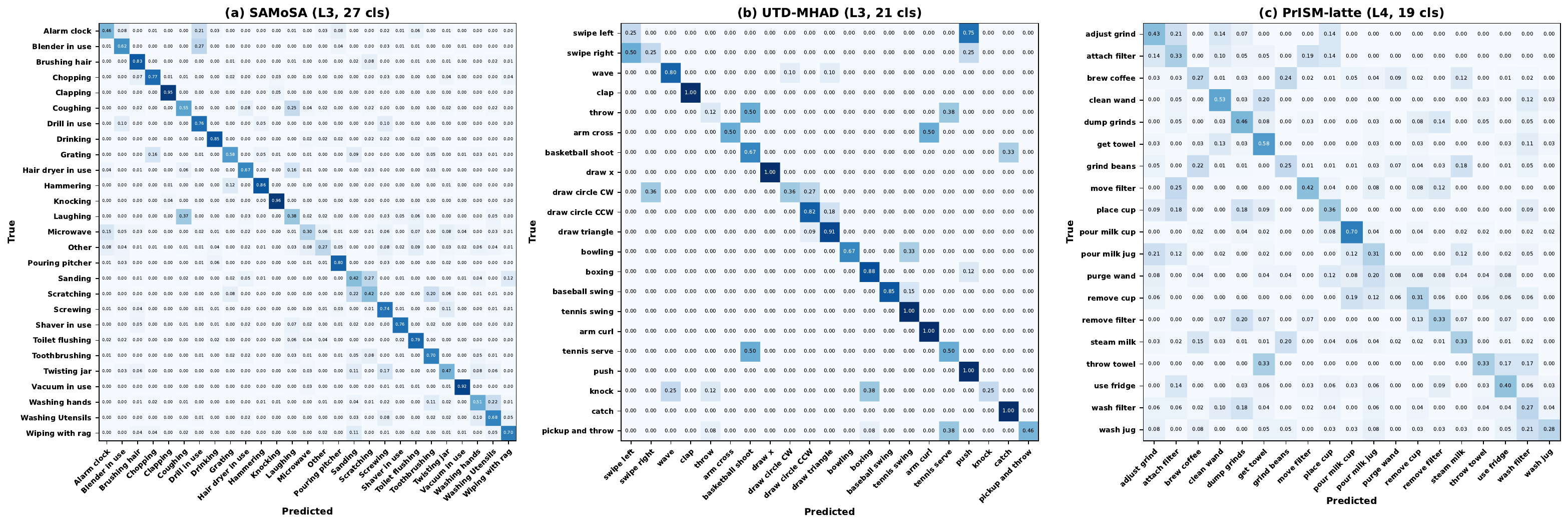}
  \caption{Row-normalized confusion matrices for the frozen 6-axis encoder with a linear probe, full split, on the three held-out benchmarks. Errors concentrate among kinematically similar classes, with PrISM latte showing the broadest off-diagonal mass.}
  \Description{Three row-normalized confusion matrices in a blue colormap, covering SAMoSA with 27 classes, UTD-MHAD with 21 classes, and PrISM-latte with 19 classes, each with true classes on the vertical axis and predicted classes on the horizontal axis. All three show a clear dark diagonal, UTD-MHAD has the cleanest diagonal with little else visible, SAMoSA shows scattered light cells near the diagonal among neighboring kitchen and workshop activities, and PrISM-latte carries the most off-diagonal mass, spread broadly across the whole grid rather than concentrated in a few pairs.}
  \label{fig:app_confusion_offline}
\end{figure*}

\begin{figure*}[h]
  \centering
  \includegraphics[width=\textwidth]{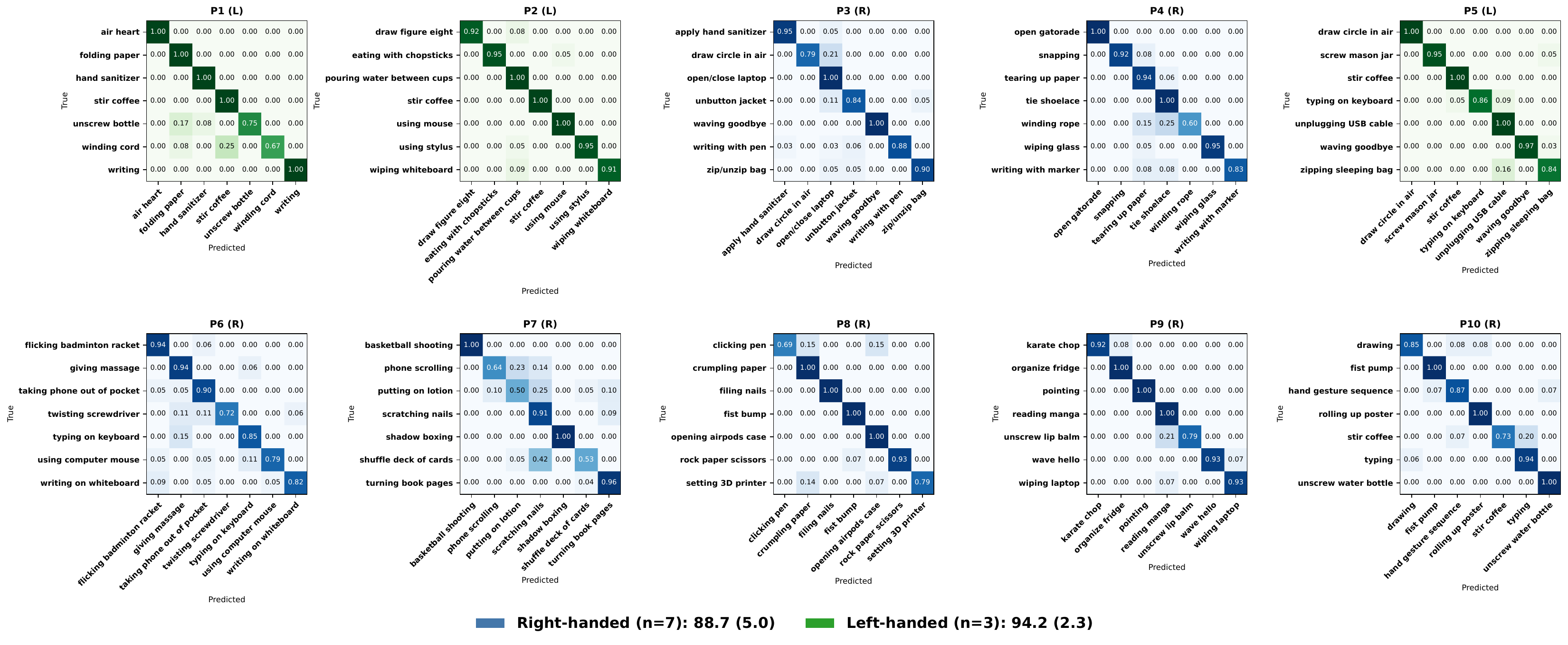}
  \caption{Per-participant confusion matrices for the 6-axis encoder in the full-session setting, colored by which wrist the watch was worn on.}
  \Description{Ten row-normalized confusion matrices arranged in two rows of five, one per participant, each titled with the participant identifier and the wrist the watch was worn on. Matrices for right-handed participants are shaded blue and those for left-handed participants green, with a legend giving group means of 88.7 percent for the seven right-handed participants and 94.2 percent for the three left-handed ones. Nearly every matrix shows a strong dark diagonal over its seven self-chosen activities, and the clearest off-diagonal cells appear for P7, where scrolling, lotion, and nail scratching trade predictions.}
  \label{fig:app_confusion_user}
\end{figure*}

Figure~\ref{fig:userstudy_hardmap} maps per-class accuracy for every participant-activity pair in the personalization study, Figure~\ref{fig:app_confusion_offline} gives per-class confusion structure on the three held-out benchmarks, and Figure~\ref{fig:app_confusion_user} gives the per-participant matrices.

\section{Activity Bank and Participant Details}
\label{app:activities}
Table~\ref{tab:activity_bank} lists the 55 candidate activities shown to participants, and Table~\ref{tab:userstudy_activity_map} gives each participant's chosen set. Of the 70 activities selected, 43 were taken directly from the bank, 12 were participant-specific variants of bank items, and 15 were new activities with no bank equivalent.

\begin{table*}[h]
\centering
\small
\caption{Activity bank shown to participants to illustrate the target granularity of on-demand wrist activity recognition. Participants could select from this list or define their own activities.}
\begin{tabularx}{\textwidth}{lX}
\toprule
\textbf{Category} & \textbf{Candidate activities shown to participants} \\
\midrule
Hand / air gestures &
drawing a heart in the air; drawing a circle in the air; drawing a figure-eight in the air; waving goodbye; conducting an orchestra; air drumming; shadow boxing; finger snapping sequence; thumbs up/down toggle; pointing at corners of the room; tracing the alphabet in the air; karate chop motion; fist pump; jazz hands; air guitar strumming \\
\midrule
Object manipulation (common lab items) &
stirring coffee with a spoon; opening/closing a laptop lid; turning a doorknob; flipping pages of a book; stacking/unstacking cups; shuffling a deck of cards; peeling tape off a roll; crumpling paper into a ball; folding a sheet of paper in half; uncapping and recapping a marker; rolling a ball of clay; winding a cord around your hand; zipping/unzipping a bag; tearing paper into strips; clicking a retractable pen repeatedly \\
\midrule
Tool-like / workspace &
typing on a keyboard; using a computer mouse (clicking and dragging); writing with a pen on paper; erasing a whiteboard; cutting paper with scissors; stapling papers; using a screwdriver; tightening a jar lid; pouring water between cups; wiping a table with a cloth; peeling a sticky note off a pad; pressing buttons on a calculator; turning a dial/knob; plugging/unplugging a USB cable; sharpening a pencil \\
\midrule
Personal / body &
brushing hair; applying hand sanitizer (rubbing hands); putting on / taking off a wristwatch; buttoning / unbuttoning a shirt cuff; tying a shoelace; rolling up a sleeve; stretching a rubber band between fingers; cracking knuckles (hand flex); filing nails with an emery board; putting on and removing a glove \\
\bottomrule
\end{tabularx}
\label{tab:activity_bank}
\end{table*}

\begin{table*}[ht]
\centering
\small
\caption{Participant-specific activity sets used in the on-demand personalization study. To preserve readability in the main figures, activities are shown as Act 1--Act 7 in participant order; this table provides the corresponding semantic labels.}
\begin{tabularx}{\textwidth}{c X}
\toprule
\textbf{Participant} & \textbf{Chosen activities} \\
\midrule
P1 & air heart; folding paper; hand sanitizer; stirring coffee; unscrewing bottle; winding cord; writing \\
P2 & drawing figure eight in the air; eating with chopsticks; pouring water between cups; stirring coffee; using mouse; using stylus; wiping whiteboard \\
P3 & applying hand sanitizer; drawing circle in the air; opening and closing laptop; unbuttoning jacket; waving goodbye; writing with pen; zipping and unzipping bag \\
P4 & opening Gatorade bottle; snapping; tearing paper; tying shoelace; winding rope; wiping glass; writing with marker \\
P5 & drawing circle in the air; screwing mason jar lid; stirring coffee; typing on keyboard; unplugging USB cable; waving goodbye; zipping sleeping bag \\
P6 & flicking badminton racket; massage giving; taking phone out of pocket; twisting screwdriver; typing on keyboard; using computer mouse; writing on whiteboard \\
P7 & basketball shooting; phone scrolling; putting on lotion; scratching nails; shadow boxing; shuffling deck of cards; turning book pages \\
P8 & clicking pen; crumpling paper; filing nails; fist bump; opening AirPods case; rock paper scissors; setting 3D printer \\
P9 & karate chop; organizing fridge; pointing; reading Japanese manga; unscrewing lip balm; waving hello; wiping laptop \\
P10 & drawing; fist pump; hand gesture sequence; rolling up poster; stirring coffee; typing; unscrewing water bottle \\
\bottomrule
\end{tabularx}
\label{tab:userstudy_activity_map}
\end{table*}

\end{document}